\documentclass[nohyperref]{article}

\usepackage{subcaption}
\usepackage{microtype}
\usepackage{graphicx}
\usepackage{booktabs} 

\usepackage{hyperref}

\usepackage[accepted]{icml2023}

\usepackage{amsmath}
\usepackage{amssymb}
\usepackage{mathtools}
\usepackage{amsthm}

\usepackage[capitalize,noabbrev]{cleveref}

\theoremstyle{plain}

\theoremstyle{definition}

\theoremstyle{remark}

\usepackage[textsize=tiny]{todonotes}

\newcommand*{\circles}{\textsc{Circles}}
\newcommand*{\windmill}{\textsc{Windmill}}
\newcommand{\ind}{\perp\!\!\!\perp} 

\icmltitlerunning{Spurious Correlations and Where to Find Them}

\begin{document}

\twocolumn[
\icmltitle{Spurious Correlations and Where to Find Them}



\icmlsetsymbol{equal}{*}

\begin{icmlauthorlist}
\icmlauthor{Gautam Sreekumar}{msu}
\icmlauthor{Vishnu Naresh Boddeti}{msu}
\end{icmlauthorlist}

\icmlaffiliation{msu}{Department of CSE, Michigan State University, East Lansing, MI-48823, USA}

\icmlcorrespondingauthor{Gautam Sreekumar}{sreekum1@msu.edu}

\icmlkeywords{spurious correlations, causal graphs}

\vskip 0.3in
]



\printAffiliationsAndNotice{}  

\begin{abstract}
Spurious correlations occur when a model learns unreliable features from the data and are a well-known drawback of data-driven learning. Although there are several algorithms proposed to mitigate it, we are yet to jointly derive the indicators of spurious correlations. As a result, the solutions built upon standalone hypotheses fail to beat simple ERM baselines. We collect some of the commonly studied hypotheses behind the occurrence of spurious correlations and investigate their influence on standard ERM baselines using synthetic datasets generated from causal graphs. Subsequently, we observe patterns connecting these hypotheses and model design choices.
\end{abstract}
\section{Introduction\label{sec:introduction}}

Spurious correlation is a well-studied problem in machine learning literature and several solutions have been proposed to mitigate it~\cite{arjovsky2019invariant}. Despite these best attempts, empirical risk minimization~(ERM)~\cite{vapnik1999nature} remains a strong baseline~\cite{gulrajani2020search}. We believe that the first step in developing robust solutions against spurious correlations is recognizing \emph{when} the models trained using ERM succumb to spurious correlations.

Several factors have been proposed as indicators of spurious correlations. These include overparameterization, partial predictiveness of invariant features, and the amount of data from different environments. However, existing studies about the occurrence of spurious correlation limit their scope to one or a few of these factors. For example, \citet{sagawa2020investigation} study the effect of overparameterization on underrepresented groups in the training data but do not investigate this phenomenon on easy-to-learn tasks. This was theoretically analyzed by \citet{nagarajan2020understanding}, although they used maximum margin models.

\begin{figure}[!ht]
    \centering
    \subcaptionbox{Observational graph\label{fig:obs-causal-graph}}[0.23\textwidth]{\includegraphics[width=0.2\textwidth]{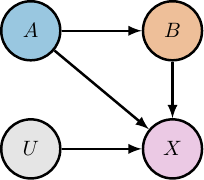}}
    \subcaptionbox{Interventional graph\label{fig:intB-causal-graph}}[0.23\textwidth]{\includegraphics[width=0.2\textwidth]{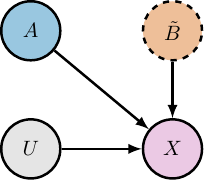}}
    \caption{Our objective is to predict the binary variables $A$ and $B$ from the observed data $X$. Causal modeling of the data generation graph helps to study spurious correlation.~(\Cref{subsec:causal-spurious-correlation})\label{fig:causal-graph}}
\end{figure}

However, in practice, these factors may not appear separately. For example, \citet{hill2021progress} noted that SARS-CoV-2 datasets were dominated by samples from the US and the UK due to sampling bias, which ``may lead to false conclusions about true transmission pathways of virus lineages". Additionally, the connectedness of air networks is a larger factor in the spread of air-borne diseases than geographical distance~\cite{lemey2014unifying}. Thus, developed countries can introduce both sampling bias and additional confounders to the dataset.

Therefore, the interaction of these factors with the model must be jointly analyzed. Our goal is to investigate how these factors affect the models trained using ERM. To that end, we develop synthetic datasets using causal modeling that allow us to finely adjust the potential factors causing spurious correlation. Knowing the causal model also facilitates relating spurious correlations to dependence relations in the causal graph. We hope that the findings of this paper serve as a guiding principle for the future development of solutions to mitigate spurious correlations.

\section{Background\label{sec:background}}

Spurious correlations occur when a model learns correlations from the observed data that do not hold under natural distribution shifts\footnote{``Natural" distribution shifts are found in passively collected data. For instance, a ``STOP" sign with chipped edges. An example of an ``unnatural" distribution shift is a green ``STOP" sign.}. A robust model is expected to use only invariant features that are reliable during testing. Thus, observed data $X$ is assumed to consist of invariant (or core) features $X_{\text{inv}}$ and spurious features $X_{\text{sp}}$. In practice, spurious features could occur due to biases during data-collection or measurement errors~\cite{fan2014challenges}. Several hypotheses exist about the origin of spurious correlations from a learning perspective.

\noindent\textbf{Partially-predictive invariant features:} The most common hypothesis about spurious correlations is that a model relies on spurious features when the invariant features are only partially-predictive of the downstream task or are less useful compared to spurious features~\cite{sagawa2020investigation}. This hypothesis has been further used to show how adversarial attacks exploit a model's dependence on spurious features~\cite{ilyas2019adversarial, zhang2021causaladv}.

\noindent\textbf{Simplicity bias:} A related hypothesis is that SGD-trained neural networks prefer to learn simple features~\cite{shah2020pitfalls, valle2018deep}. As a result, these models may rely on simpler, but less predictive spurious features instead of sophisticated, yet fully predictive invariant features.

\noindent\textbf{Majority advantage:} Another factor that may result in spurious correlations is statistical bias~\cite{nagarajan2020understanding}. If spurious features are present in the majority of the data samples, the model would rely on them to minimize the training error. Mitigation tools derived from this hypothesis usually exploit the diversity in training data to ensure that the model sees enough examples without these spurious features during training.~\cite{arjovsky2019invariant, wang2021identifying, idrissi2022simple}.

\noindent\textbf{Other hypotheses} that have been studied include noisy invariant features~\cite{khani2020feature}, imperfect partitions of the training data that allow group-specific spurious correlations~\cite{zhou2021examining} and invariant features given less weight by the final layers~\cite{kirichenko2022last, izmailov2022feature}.

In our experiments, we consider datasets in which the invariant features vary in their predictive power, proportion in the training data, and complexity.

\begin{figure*}[!ht]
    \centering
    \includegraphics[width=0.4\textwidth]{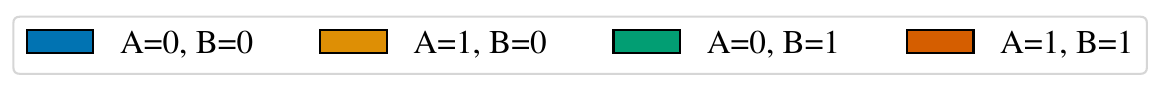} \\
    \subcaptionbox{\circles{} dataset\label{fig:circles-dataset}}{\includegraphics[width=0.3\textwidth]{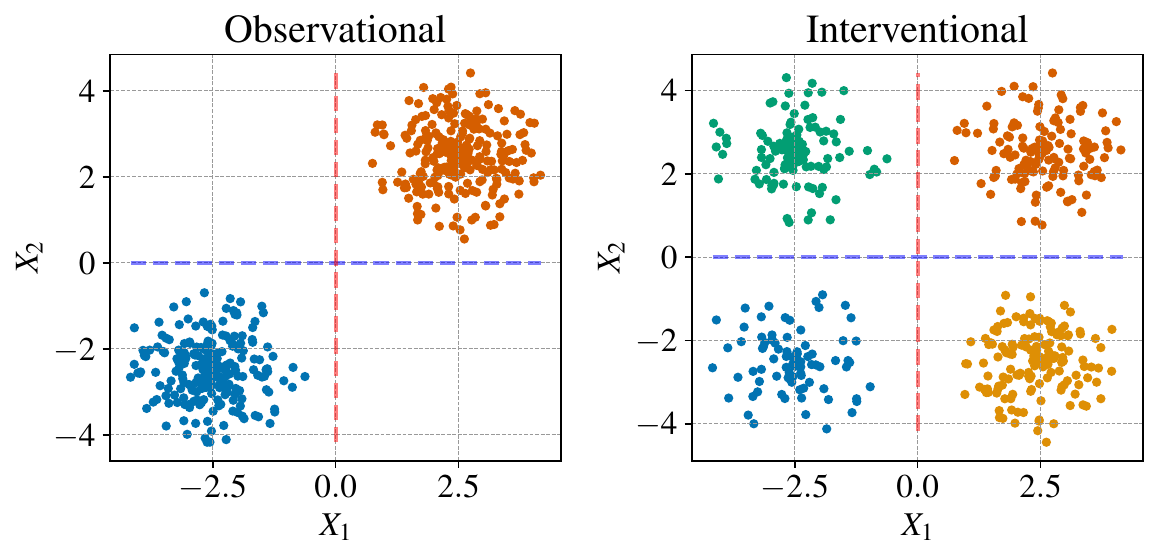}}
    \subcaptionbox{Impossible \circles{} dataset\label{fig:impossible-circles-dataset}}{\includegraphics[width=0.3\textwidth]{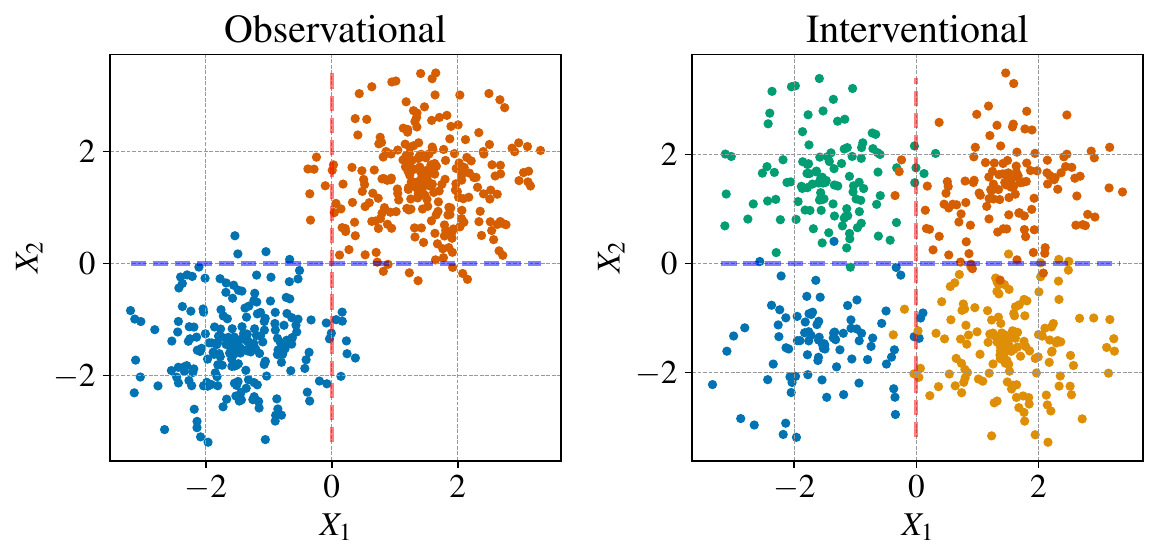}}
    \subcaptionbox{\windmill{} dataset\label{fig:windmill-dataset}}{\includegraphics[width=0.3\textwidth]{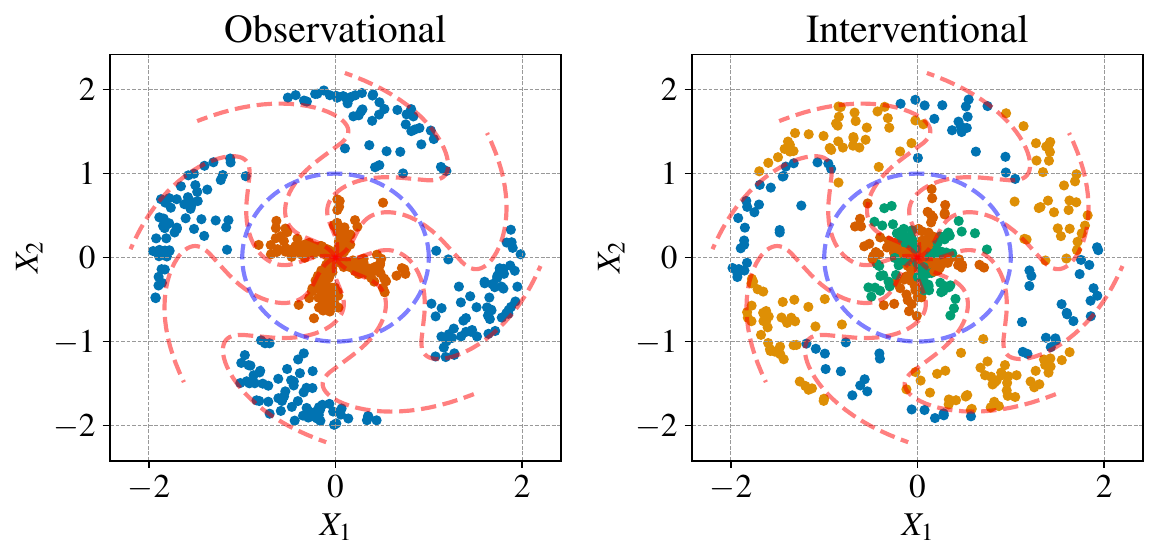}}
    \caption{Synthetic datasets constructed to study the occurrence of spurious correlations. \textcolor{red}{Red} and \textcolor{blue}{blue} dotted lines indicate the optimal decision boundaries of $A$ and $B$ respectively.\label{fig:dataset-samples}}
\end{figure*}
\section{Setup\label{sec:setup}}

\subsection{Causal interpretation of spurious correlation\label{subsec:causal-spurious-correlation}}

Spurious correlations are often due to confounders and unobserved correlated variables, and not due to true causal relations. As a result, they are affected by natural distribution shifts. Therefore, causal graphs are a suitable choice to model the occurrence of spurious correlations. By treating prediction tasks as anti-causal learning, we can model the observed data $X$ as $X=f_X(U_X, Y_1, Y_2, \dots, Y_n)$ where $Y_1, Y_2, \dots, Y_n$ are label variables and $U_X$ is an exogenous variable denoting unobserved factors that affect $X$. The label variables may be causally related to each other as $Y_i=f_{Y_i}(\textbf{Pa}(Y_i), U_{Y_i})$ where $\textbf{Pa}(Y_i)$ denote the parent variables of $Y_i$ in the causal graph and $U_{Y_i}$ denote the unobserved factors affecting $Y_i$.

For the model to distinguish spurious features from invariant features, the training set must contain samples where spurious correlations do not hold. In~\cite{arjovsky2019invariant}, samples were collected from different environments to break spurious correlations. Since we model our data-generating process as a causal graph, different environments correspond to different distributions of the label variables. One way to induce different distributions is through interventions on label variables. By intervening on, say, $Y_i$, it becomes independent of its parent variables $\textbf{Pa}(Y_i)$. Let $F_{Y_i}$ be the feature learned by a model to predict $Y_i$. If $F_{Y_i}$ has spurious information, it may have non-zero dependence with some variable $Y_j\in \textbf{Pa}(Y_i)$ during interventions.

\subsection{Datasets\label{subsec:datasets}}

For our study, we consider the simplest causal graph with two binary variables $A$ and $B$. The causal model of the data-generation process is shown in~\Cref{fig:obs-causal-graph}. Our task is to predict the binary labels from the observed data $X$. Note that $X$ embodies both the invariant feature $X_{\text{inv}}$ and the spurious feature $X_{\text{sp}}$. Although invariant and spurious features are treated as separate casual variables in existing works~\cite{khani2021removing, arjovsky2019invariant}, we model them jointly since they may be entangled in practice. The unobserved factors of variation are collectively denoted by $U$. Since $A$ is a parent of $B$, a change in distribution of $A$ affects that of $B$. However, vice versa does not hold. In our experiments, the potential spurious correlation that a model may learn is to use features corresponding to $B$ to predict $A$.

We construct two synthetic datasets -- (1)~\circles{} and (2)~\windmill{}. For each dataset, we collect observational and interventional data points following~\Cref{fig:obs-causal-graph} and~\Cref{fig:intB-causal-graph} respectively. The exact functional formulations of the datasets are provided in~\Cref{appsec:dataset-generation}.

\noindent\textbf{\circles{} dataset:} The \circles{} dataset consists of vectors sampled from four circular regions in the $\mathbb{R}^2$-space~(\Cref{fig:circles-dataset}). Each cluster corresponds to $(A=a,B=b)$ for some $a,b\in\{0, 1\}$. $A$ and $B$ decide $X_1$ and $X_2$ in~\Cref{fig:circles-dataset} respectively. When the clusters are well-separated, a linear model can easily achieve zero test error. However, when they overlap, it is impossible to find a zero-error decision boundary~(\Cref{fig:impossible-circles-dataset}). We use the \circles{} dataset to analyze spurious correlations in easy-to-learn and impossible-to-learn tasks.

\noindent\textbf{\windmill{} dataset:} Our second dataset is designed to explicitly prompt spurious correlations. The effects of the variables $A$ and $B$ on the observed data $X$ are entangled and the true decision boundary for $A$ is more difficult to learn than that of $B$\footnote{``Difficulty to learn" is measured in terms of the minimum degree required by a polynomial to approximate it with zero test error.}. The complexity of the true decision boundary of $A$ is adjusted through a parameter $\lambda_{\text{off}}$. Higher the value of $\lambda_{\text{off}}$, the higher the complexity is. As predicting $A$ using its invariant features alone becomes more difficult, the model tends to use spurious features. Observational and interventional data samples from \windmill{} dataset are illustrated in~\Cref{fig:windmill-dataset}.

\subsection{Why synthetic data?\label{subsec:why-synthetic}}

The key advantage of synthetic datasets over natural ones is the ability to adjust their properties precisely. Specifically, the \windmill{} dataset allows us to change the complexity of its decision boundaries, and the \circles{} dataset allows us to alter the geometric coordinates and dimensions of the clusters. Synthetic datasets have been used in prior works to analyze the evolution of features learned by models~\cite{hermann2020shapes}.
\section{Experiments\label{sec:experiments}}

We design experiments to study the impact of task difficulty, statistical bias, and predictive power of invariant features. We first distinguish tasks based on their difficulty -- easy-to-learn, difficult-to-learn, and impossible-to-learn. We then vary the amount of interventional data in each dataset. Then for each such dataset, we vary the capacities of the models by changing the depth and the width of the MLPs.

\noindent\textbf{Method:} We do not propose any novel method to mitigate spurious correlation. Our objective is to unify possible factors that contribute to spurious correlation in models trained using ERM. We consider two commonly followed sub-paradigms under ERM -- (1)~standard ERM~(simply referred to as ERM), (2)~ERM with resampling~(ERM-Resampled). A single training batch in ERM comprises both observational and interventional samples. In contrast, observational and interventional samples never appear in the same batch in ERM-Resampled. Existing works~\cite{idrissi2022simple, gulrajani2020search} indicate that ERM-Resampled is a strong baseline against spurious correlation and in domain generalization. Throughout our experiments, we use MLPs with ReLU as the activation function.

\noindent\textbf{Metrics:} The standard sign of spurious correlations is a drop in test accuracy due to a change in distribution. Therefore, we quantify spurious correlations in the model using the relative drop in test accuracy between observational and interventional samples. Additionally, since we know that the intervened variable must be independent of its parents, we evaluate the robustness of the features by measuring the dependence between the features on interventional samples.

\noindent\textbf{Measuring dependence:} Several methods have been proposed to measure dependence between high-dimensional vectors, of which kernel-based methods are popular~\cite{hsic, kcc}. However, these are difficult to interpret from their absolute values. Therefore, we devise a new independence measure based on statistical independence testing called ``ratio over independent samples"~(RoIS). Given two features $F_A$ and $F_B$ with $N$ samples, RoIS is measured as $\text{RoIS}(F_A, F_B) = \frac{\text{dep} \left( F_A, F_B \right)}{\frac{1}{K} \sum_{i=1}^{K} \text{dep}\left( F_A^{(\pi_i)}, F_B \right)}$, where $\pi_i$ is some permutation of $N$ samples and $\text{dep}$ is our choice of measure of dependence. In our experiments, we use a normalized version of HSIC~\cite{hsic} as $\text{dep}$. Refer to~\Cref{app:rois} for a detailed description.
\begin{figure*}[!ht]
    \centering
    \subcaptionbox{Relative drop in test accuracy\label{fig:windmill-model-capacity-difficulty-dropacc}}{\includegraphics[width=0.47\textwidth]{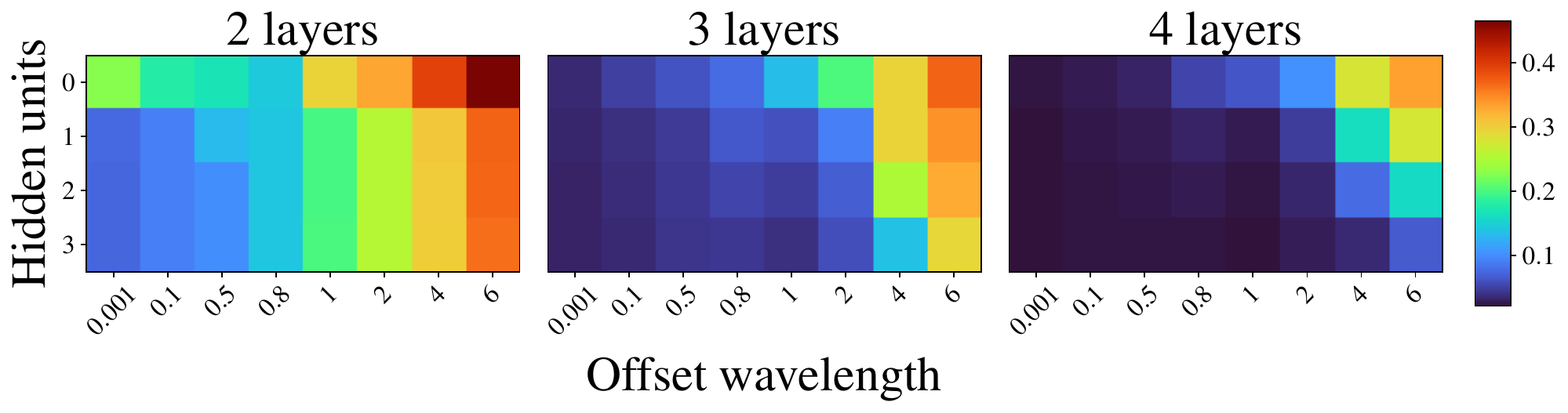}}
    \subcaptionbox{Dependence between features $\text{RoIS}(F_A,F_B)$\label{fig:windmill-model-capacity-difficulty-rois}}{\includegraphics[width=0.47\textwidth]{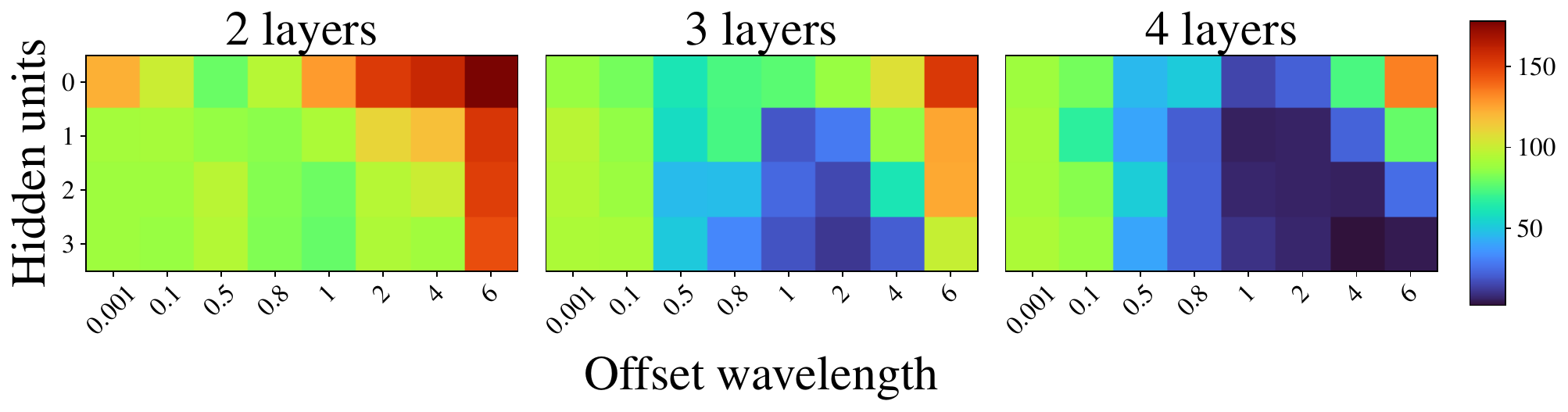}}
    \caption{Effect of task complexity on ERM-Resampled models for difficult-to-learn tasks\label{fig:windmill-model-capacity-difficulty}}
\end{figure*}

\subsection{Easy-to-learn tasks\label{subsec:easy-to-learn}}

We construct easy-to-learn tasks similar to those proposed in \cite{nagarajan2020understanding} using our \circles{} dataset. The true decision boundaries for $A$ and $B$ are linear in all cases, \textit{i.e.}, a single non-trivial parameter is sufficient to learn the true boundary for each label. We train MLPs for each task through ERM-Resampled and measure the relative drop accuracy between observational and interventional data during testing.

\subsubsection{Effect of amount of interventional data\label{subsubsec:beta-easy-circles}}

Each dataset contains $N$ samples: $\beta N$ observational and $(1-\beta)N$ interventional, where $0<\beta<1$. Varying $\beta$ creates what is referred to as ``statistical skew" in~\cite{nagarajan2020understanding}.

\begin{figure}[!ht]
    \centering
    \subcaptionbox{Due to change in $\beta$\label{fig:circle-model-capacity-beta}}{\includegraphics[width=0.23\textwidth]{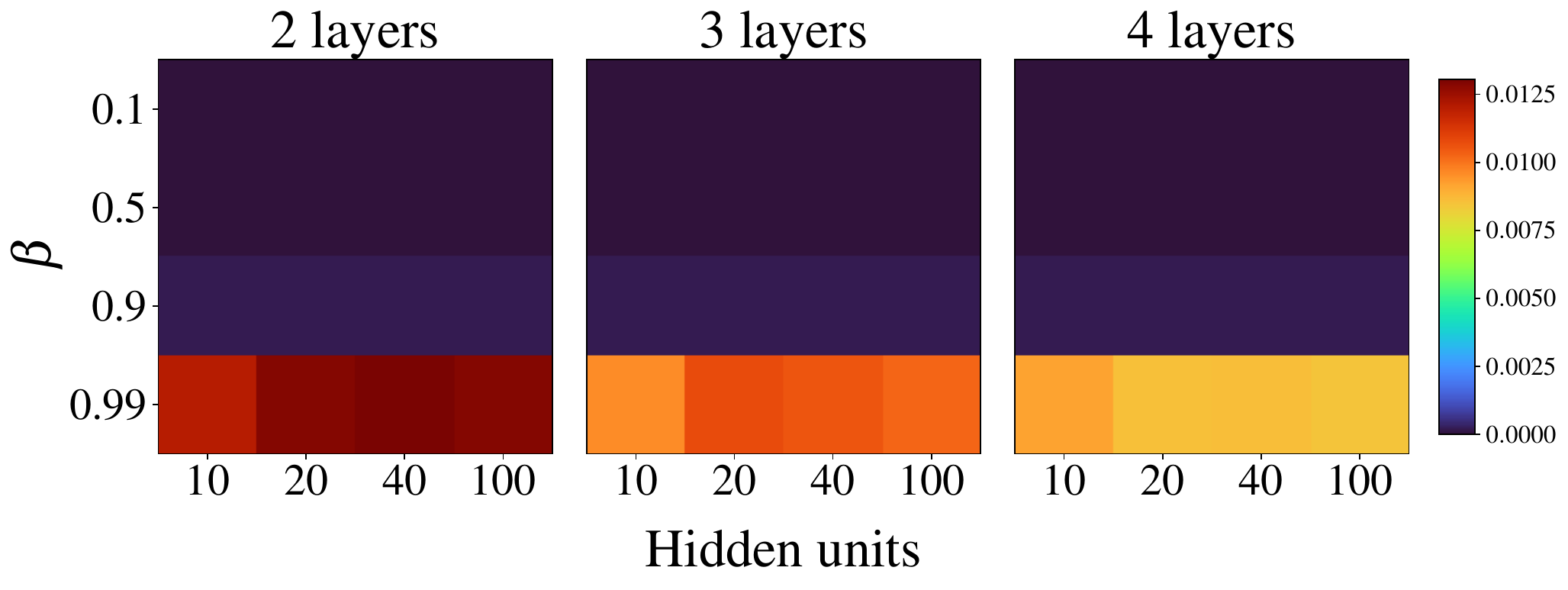}}
    \subcaptionbox{Due to change in radius ratio\label{fig:circle-model-capacity-ratio}}{\includegraphics[width=0.23\textwidth]{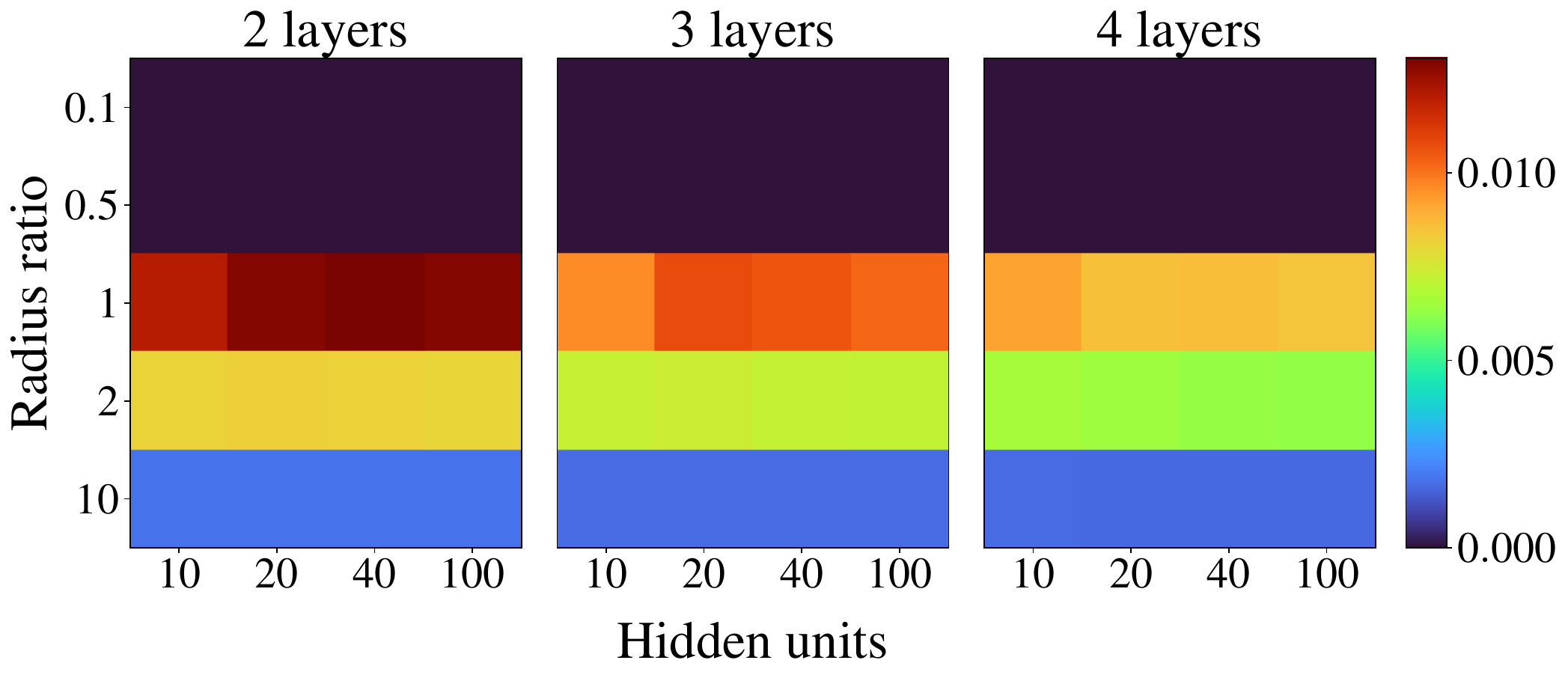}}
    \caption{Relative drop in test accuracy of ERM models in easy-to-learn tasks}
\end{figure}

\noindent\textbf{Analysis:} We trained models using the ERM-Resampled scheme. Surprisingly, we found that there was no drop in accuracy due to a variation of $\beta$. Even with just 1\% interventional data, the model learned the true decision boundary. We attribute this to ERM-Resampled being competitively robust to spurious correlations. We do not report these results since they are trivial. Instead, we consider standard ERM, which is a weaker baseline. We observe that even a weak baseline like ERM is robust to spurious correlations until $\beta$ reaches around 0.99. \Cref{fig:circle-model-capacity-beta} visualizes the relative accuracy drop for models trained using standard ERM.

\subsubsection{Effect of geometric distortion\label{subsubsec:ratio-easy-circles}}

We now consider the effect of the cluster shapes on spurious correlation. We adjust the ratio between the radii along the horizontal and vertical dimensions axis in our \circles{} dataset and observe the occurrence of spurious correlation. The ratio can be written as $\frac{r_B}{r_A}$ where $r_A$ and $r_B$ are the radii along horizontal and vertical directions respectively. Changing the radius ratio essentially shears each circular cluster into an ellipse. Since we found ERM susceptible to spurious correlation only at larger values of $\beta$, we set $\beta=0.99$ during this experiment.

\noindent\textbf{Analysis:} \Cref{fig:circle-model-capacity-ratio} shows the drop in test accuracy due to the change in radius ratio. We observe the drop in test accuracy is~(1) very small ($< 1\%$), and~(2) limited to radius ratio $\geq 1$, \textit{i.e.}, only when the circular clusters are vertically sheared. This could be due to horizontal shearing limiting the number of decision boundaries that achieve zero training error, especially when the number of interventional samples is low. Furthermore, as the shearing increased along the vertical dimension, the drop in test accuracy decreased. Refer to~\Cref{app:easy-circles-ratio-decision-boundaries} for the visualization of corresponding decision boundaries.

\subsection{Difficult-to-learn tasks\label{subsec:difficult-to-learn}}

We now design a task where spurious features are easier to be learned compared to invariant features. The phenomenon is commonly referred to as ``simplicity bias". To investigate the occurrence of spurious correlation due to simplicity bias, we design tasks where the model has to learn a complex, but zero-test error decision boundary from the training data. Here, the ``complexity" of a decision boundary can be roughly defined as the minimum degree required by a polynomial to fully approximate it. \Cref{fig:windmill-difficulty} illustrates different complexity levels of \windmill{} dataset due to change in $\lambda_{\text{off}}$ in \Cref{appsec:windmill-generation}.
\begin{figure}[!ht]
    \centering
    \subcaptionbox{Relative drop in test accuracy\label{fig:windmill-model-capacity-beta-dropacc}}{\includegraphics[width=0.24\textwidth]{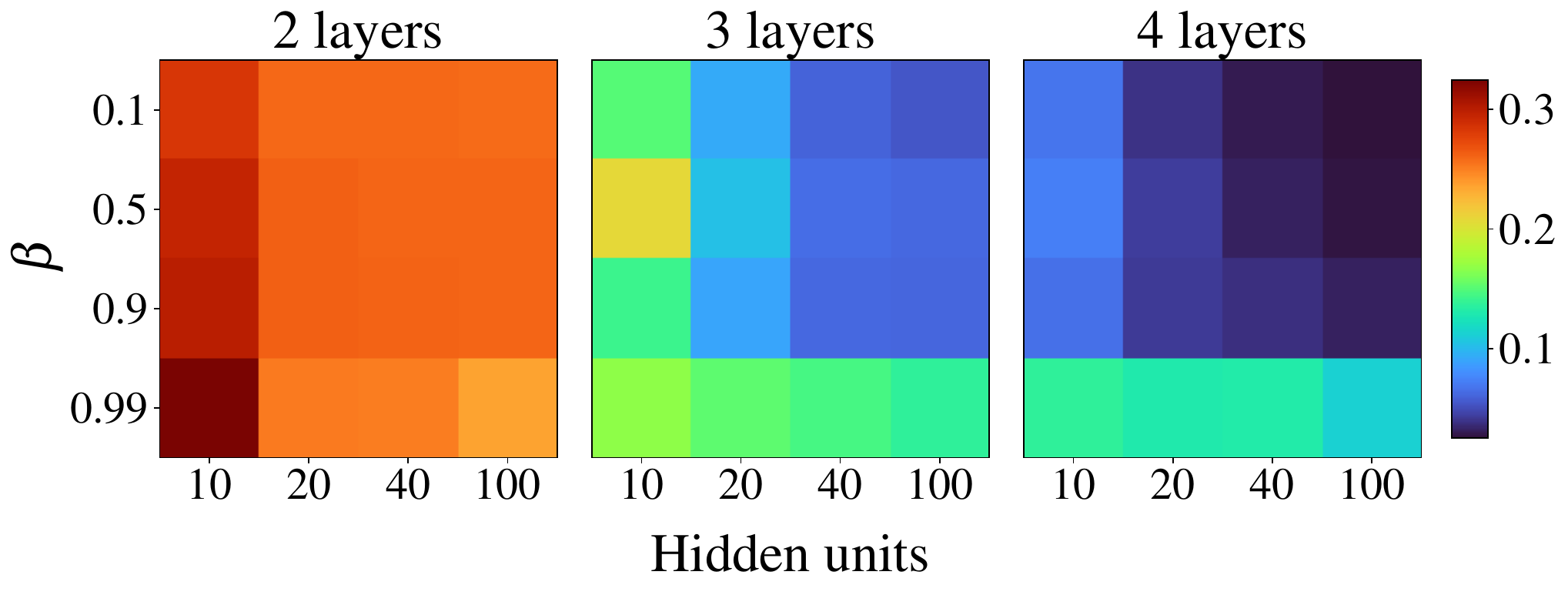}}
    \subcaptionbox{Dependence between features $\text{RoIS}(F_A,F_B)$\label{fig:windmill-model-capacity-beta-rois}}{\includegraphics[width=0.24\textwidth]{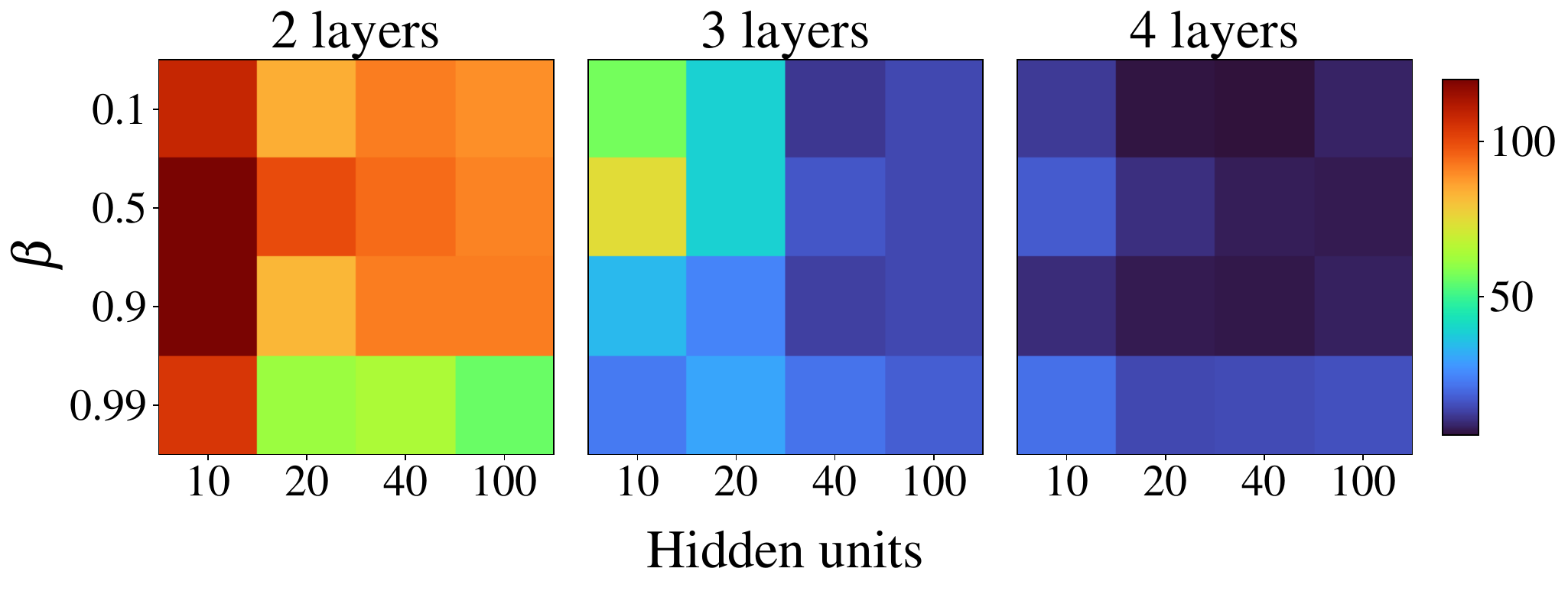}}
    \caption{Effect of amount of interventional data on ERM-Resampled models for difficult-to-learn tasks\label{fig:windmill-model-capacity-beta}}
\end{figure}
\vspace*{-0.4cm}

\subsubsection{Effect of variation of task difficulty\label{subsubsec:windmill-difficulty-variation}}

Our \windmill{} dataset allows us to adjust the complexity of the decision boundary of $A$ by varying $\lambda_{\text{off}}$. We hypothesize that a model with fixed capacity will increasingly depend on spurious features to make predictions as the complexity of the task increases.

\noindent\textbf{Analysis:} In~\Cref{fig:windmill-model-capacity-difficulty-dropacc}, we vary the complexity of the task via $\lambda_{\text{off}}$, and calculate the relative drop in accuracy between observational and interventional data for a model trained using ERM-Resampled. We make the following observations: (1)~For a fixed task difficulty, the model becomes more robust as the capacity of the model increases, (2)~With a fixed capacity, the model tends to learn spurious correlations as the task complexity increases.

For additional analysis, we measure the dependency between the features using our proposed RoIS score. A lower RoIS score indicates that learned representations contain fewer spurious features. Plotting RoIS values in~\Cref{fig:windmill-model-capacity-difficulty-rois} allows us to make two observations: (1)~Spurious correlation is always accompanied by a strong dependency between the features, (2)~However, strong dependency between the features does not imply spurious correlations, as evident when the task complexity is low.

\subsubsection{Effect of amount of interventional data\label{subsubsec:windmill-beta-variation}}

Similar to~\Cref{subsubsec:beta-easy-circles}, we vary the amount of interventional data in difficult-to-learn tasks. Following the same convention, $\beta$ denotes the proportion of observational data points. We fix the task complexity by setting $\lambda_{\text{off}}=2$.

\noindent\textbf{Analysis:} \Cref{fig:windmill-model-capacity-beta} visualizes the relative drop in test accuracy and RoIS between the features due to change in $\beta$. We make the following observations: (1)~As expected, spurious correlations reduce with an increase in the amount of interventional data, irrespective of the model capacity, (2)~For a given amount of interventional data, larger models seem to be robust against spurious correlations, (3)~The variation in dependency seems to be indicative of spurious correlation, unlike in the previous case of varying task complexity.

\subsection{Impossible-to-learn\label{subsec:impossible-to-learn}}

Our final experiment tests the hypothesis about the predictive power of invariant features. We modify the \circles{} dataset such that the circular regions partially overlap~(\Cref{fig:impossible-circles-dataset}). Due to this overlap, invariant features will no longer be able to provide a zero-error decision boundary, while spurious features can during training. Such situations may occur due to high noise in the invariant features.

\begin{figure}[!ht]
    \centering
    \subcaptionbox{Relative drop in test accuracy\label{fig:impossible-model-capacity-beta-dropacc}}{\includegraphics[width=0.3\textwidth]{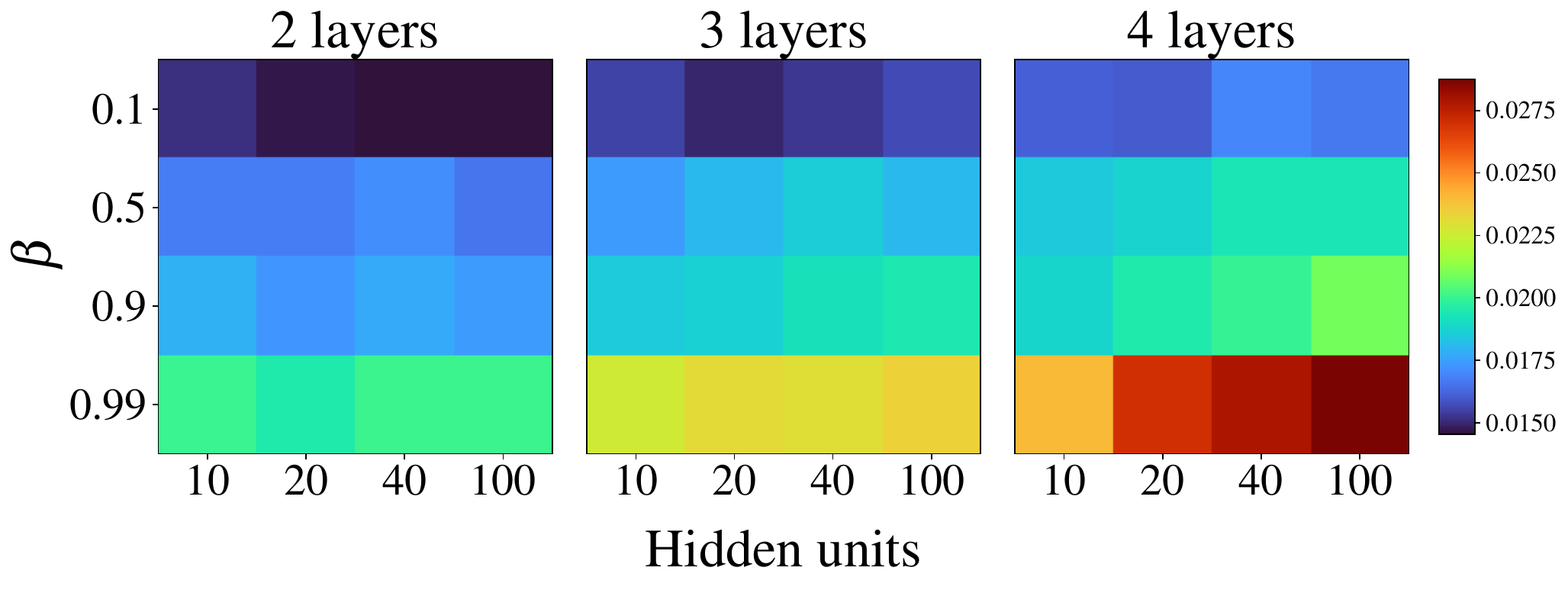}}
    \subcaptionbox{Dependence between features $\text{RoIS}(F_A,F_B)$\label{fig:impossible-model-capacity-beta-rois}}{\includegraphics[width=0.3\textwidth]{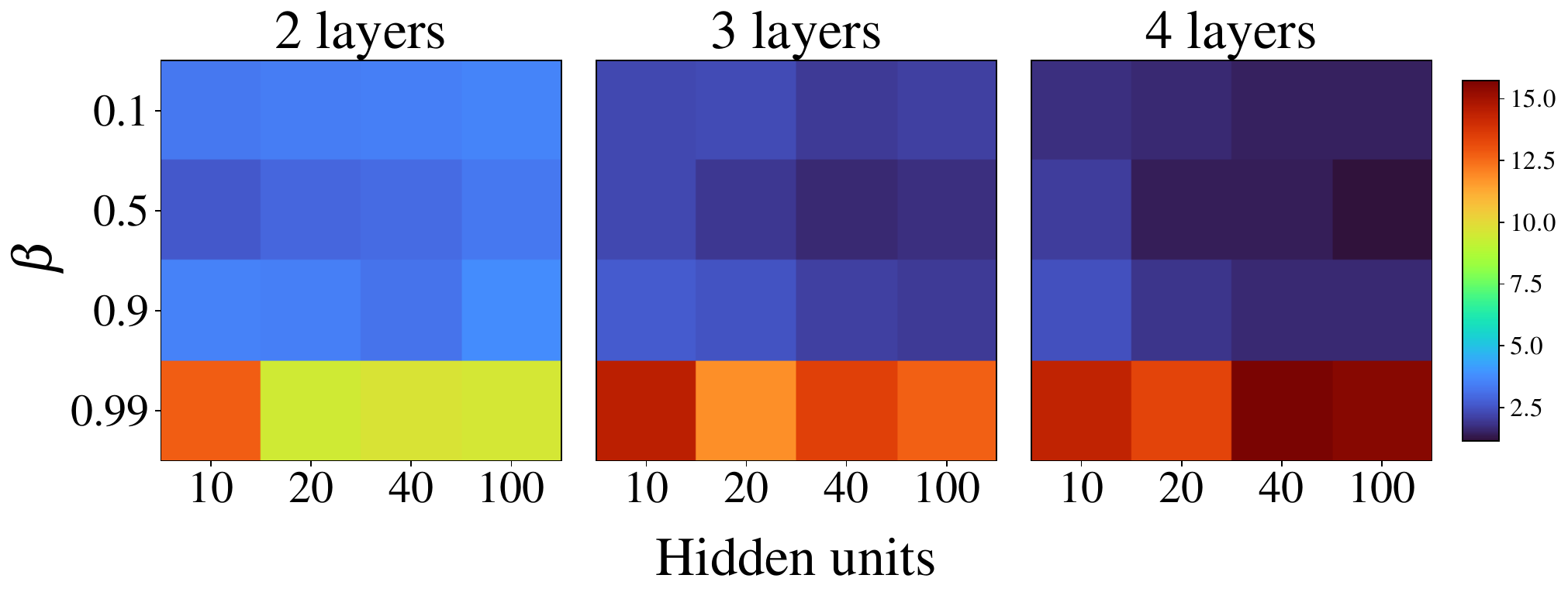}}
    \caption{Effect of amount of interventional data on ERM-Resampled models for impossible-to-learn tasks\label{fig:impossible-model-capacity-beta}}
\end{figure}

\subsubsection{Effect of amount of interventional data\label{subsubsec:beta-impossible-circles}}

Similar to our previous experiments, we vary $\beta$ and compare the relative drop in test accuracy for various models.

\noindent\textbf{Analysis: }When it is impossible to learn a zero-test error decision boundary, the behavior of the model can vary due to the data it sees, and is especially consequential when the interventional samples are too few. This is evident in \Cref{fig:impossible-model-capacity-beta} and allows us to make two interesting observations: (1)~The model becomes more robust as more interventional data is available, (2)~In contrast to difficult-to-learn tasks, larger models are more prone to spurious correlation especially when the amount of interventional data is limited, (3)~Dependency between features seems to be positively correlated with accuracy drop for lower amounts of interventional while being negatively correlated everywhere else.
\section{Concluding Remarks\label{sec:conclusion}}

Our work attempted to unify some of the commonly pursued hypotheses behind spurious correlations -- statistical bias, simplicity bias, and predictive power of the invariant features. We designed synthetic datasets that reflected these qualities following a simple causal graph. We measured the drop in test accuracy to quantify the spurious correlations learned by models trained using variants of ERM. Furthermore, our causal formulation allowed us to measure the dependence between features from interventional data.

Some of our findings were surprising. We found that SGD-based solutions were robust against spurious correlations in easy-to-learn tasks, unlike maximum margin solutions~\cite{nagarajan2020understanding}. We also found that larger models were more robust than smaller models in difficult-to-learn tasks, while smaller models were more robust in impossible prediction tasks. However, other findings were similar to those reported previously. We noted that having more interventional data improved the robustness of the model. Additionally, we showed that models learn shortcuts from data~\cite{geirhos2020shortcut} when the task is disproportionately difficult for the capacity of the model.

Our findings indicate that spurious correlations are a product of the dataset, the model, and the training scheme. Therefore, future algorithms that are proposed to mitigate spurious correlations must evaluate using models with different capacities with varying proportions of interventional points. In addition to drop in test accuracy, independence relations between known causal variables can provide a deeper understanding of the algorithms.

\bibliography{references}
\bibliographystyle{icml2023}

\newpage
\appendix
\onecolumn
\section{Dataset Generation\label{appsec:dataset-generation}}

\noindent\textbf{Notation:} $\text{Bern}(p)$ denotes a Bernoulli distribution with parameter $p$. $\mathcal{U}(a, b)$ denote a uniform distribution between $a$ and $b$. $\mathcal{C}(S)$ denote uniform categorical distribution over the elements of set $S$. $\mathcal{B}(\alpha, \beta)$ denotes a beta distribution with parameters $\alpha$ and $\beta$. Variables in uppercase denote random variables and those in lowercase denote scalar constants.

The structural causal model~(SCM)~\cite{peters2017elements} corresponding to the observational causal graph shown in~\Cref{fig:obs-causal-graph} is as follows. $f_X$ and $U$ vary with the dataset.
\begin{align*}
    A &\sim \text{Bern}(p) \\
    B &= A \tag{During observation, or} \\
    B &= \Tilde{B} \tag{During intervention} \\
    X &= f_X(A, B, U)
\end{align*}

\subsection{\circles{} dataset generation\label{appsec:circles-generation}}

The parameters are:
\begin{table}[!ht]
    \centering
    \begin{tabular}{l|l|cc}
    \toprule
    Parameter & Description & \multicolumn{2}{c}{Default value} \\
    & & Easy & Impossible \\
    \midrule
    $r^{\text{(max)}}_1$ & Maximum radius along $X_1$ direction & 2 & 2 \\
    $r^{\text{(max)}}_2$ & Maximum radius along $X_2$ direction & 2 & 2 \\
    $\mu_1$ & Shift in center of ellipse along $X_1$ direction & 2.5 & 1.5 \\
    $\mu_2$ & Shift in center of ellipse along $X_2$ direction & 2.5 & 1.5 \\
    \bottomrule
    \end{tabular}
    \caption{Parameters used for generating the \circles{} dataset, what they mean, and their default values if applicable.}
    \label{tab:circles-parameters}
\end{table}
\begin{align*}
    \Theta &\sim \mathcal{U}(0, 2\pi) \tag{Sample polar angle} \\
    R_1 &\sim \mathcal{U}(0, r^{\text{(max)}}_1) \tag{Sample polar distance along $X_1$ direction} \\
    R_2 &\sim \mathcal{U}(0, r^{\text{(max)}}_2) \tag{Sample polar distance along $X_2$ direction} \\
    R &= \frac{R_1R_2}{\sqrt{R_1^2\cos^2\Theta + R_2^2\sin^2\Theta}} \tag{Polar form of an ellipse} \\
    X_1 &= (2A-1)\mu_1 + R\cos\Theta \tag{Shift according to value of $A$} \\
    X_2 &= (2B-1)\mu_2 + R\sin\Theta \tag{Shift according to value of $B$} \\
    X &= \begin{bmatrix}X_1\\X_2\end{bmatrix}
\end{align*}

\subsection{\windmill{} dataset generation\label{appsec:windmill-generation}}

The parameters are:
\begin{table}[!ht]
    \centering
    \begin{tabular}{l|l|c}
    \toprule
    Parameter & Description & Default value \\
    \midrule
    $n_{\text{arms}}$ & Number of "arms" in \windmill{} dataset & 4 \\
    $r_{\text{max}}$ & Radius of the circular region spanned by the observed data & 2 \\
    $\theta_{\text{wid}}$ & Angular width of each arm & $\frac{0.9\pi}{n_{\text{arms}}}=0.7068$ \\
    $\lambda_{\text{off}}$ & Offset wavelength. Determines the complexity of the dataset & - \\
    $\theta_{\text{max-off}}$ & Maximum offset for the angle & $\pi/6$\\
    \bottomrule
    \end{tabular}
    \caption{Parameters used for generating \windmill{} dataset, what they mean, and their default values if applicable.}
    \label{tab:windmill-parameters}
\end{table}
\begin{align*}
    R_{B} &\sim \mathcal{B}(1, 2.5) \tag{Sample radius} \\
    R &= \frac{r_{\text{max}}}{2}\left(BR_B + (1-B)(2-R_B)\right) \tag{Modify sampled radius based on $B$} \\
    \Theta_A &\sim \mathcal{C}\left(\left\{2\pi\frac{i}{n_{\text{arms}}+1}: i = 0,\dots, n_{\text{arms}}-1\right\}\right) \tag{Choose an arm} \\
    U &\sim \mathcal{U}(0, 1) \tag{To choose a random angle} \\
    \Theta_{\text{off}} &= \theta_{\text{max-off}}\sin\left(\pi\lambda_{\text{off}}\frac{R}{r_{\text{max}}}\right) \tag{Calculate radial offset for the angle} \\
    \Theta &= \theta_{\text{wid}}\left(U-0.5\right) + A\left(\Theta_A+\frac{\pi}{n_{\text{arms}}}\right) + (1-A)\Theta_A + \Theta_{\text{off}} \tag{Angle is decided by $A$ and the radial offset} \\
    X_1 &= R\cos\Theta \tag{Convert to Cartesian coordinates} \\
    X_2 &= R\sin\Theta \\
    X &= \begin{bmatrix}X_1\\X_2\end{bmatrix}
\end{align*}

\begin{figure}[!ht]
    \centering
    \includegraphics[width=0.4\textwidth]{figs/datasets/dataset_legend.pdf} \\
    \includegraphics[width=0.116\textwidth]{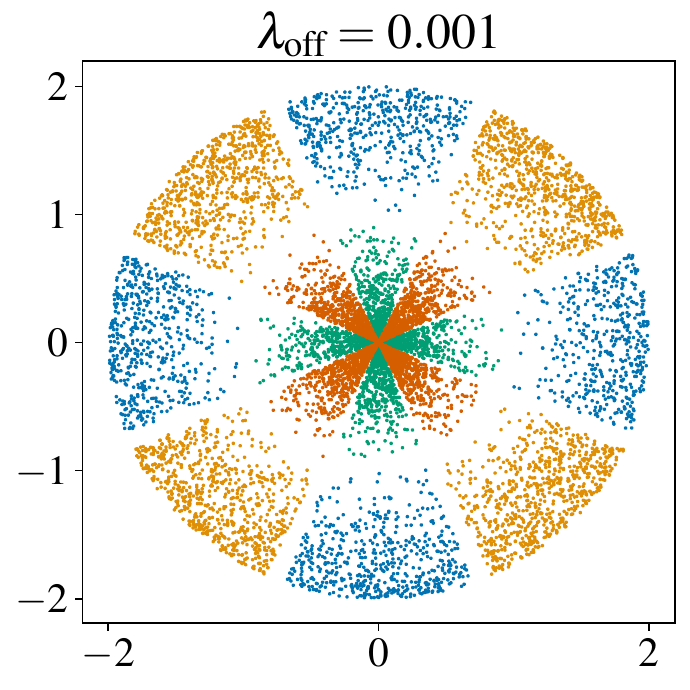}
    \includegraphics[width=0.116\textwidth]{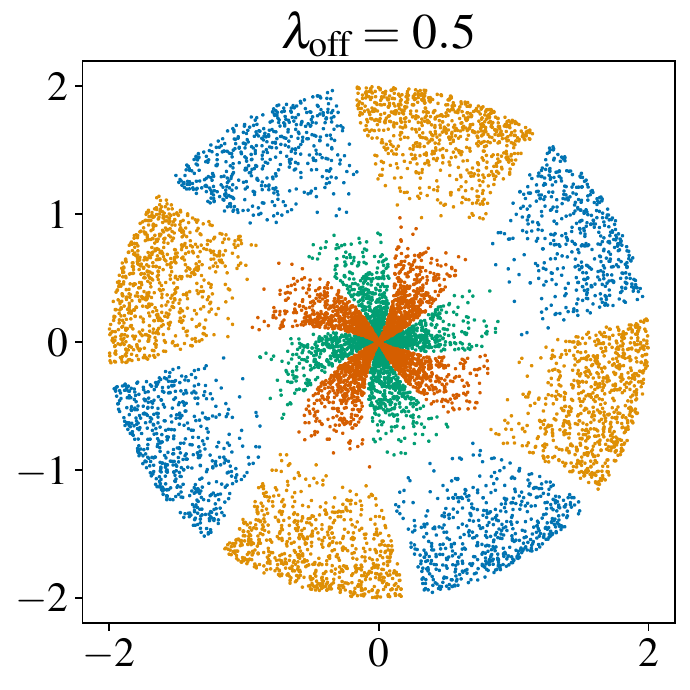}
    \includegraphics[width=0.116\textwidth]{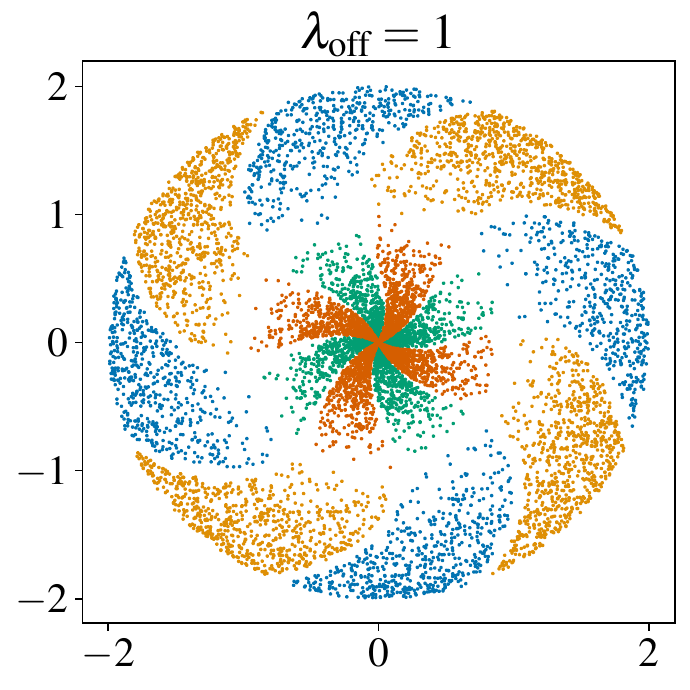}
    \includegraphics[width=0.116\textwidth]{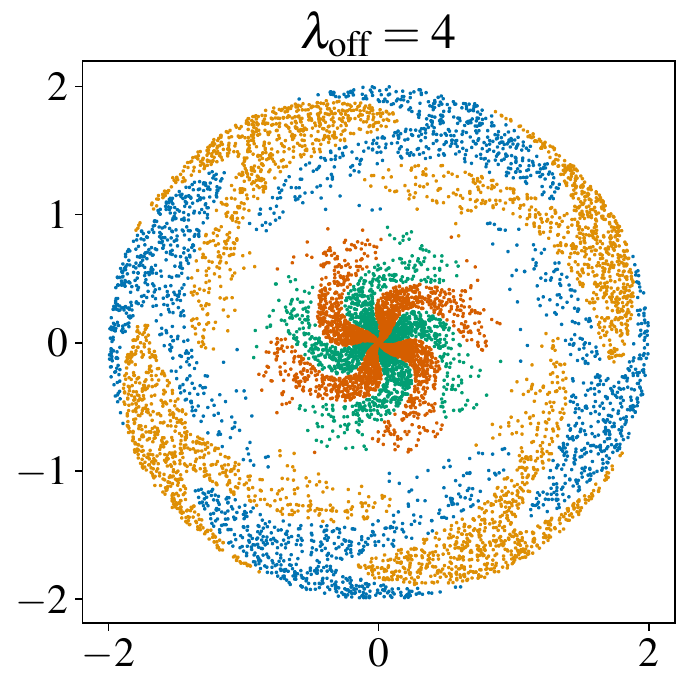}
    \caption{Difference in dataset generated by varying $\lambda_{\text{off}}$}
    \label{fig:windmill-difficulty}
\end{figure}

\Cref{fig:windmill-difficulty} shows the various datasets generated by varying $\lambda_{\text{off}}$.

\section{Quantifying dependence between features\label{app:rois}}

Suppose we wish to measure the dependence between two random vectors $X$ and $Y$ using some kernel-based measure of dependence $D(X, Y)$. Let $X$ and $Y$ consist of $n$ samples and $\{\mathcal{P}_i: i=1,\dots m\}$ be $m$ permutations of these samples. We can safely assume that $X\ind Y^{(\mathcal{P}_i)}$ for all $\mathcal{P}_i$. Therefore, the average score from the measure of dependence for independent samples can be written as,
\begin{equation}
    d^* = \frac{\sum_i^m D(X, Y^{\mathcal{P}_i})}{m}
\end{equation}
$d^*$ can be interpreted as the highest value of the measure of dependence $D$ that it may give for any pair of independent random vectors. We use $d^*$ to define our metric ``ratio over independent samples"~(RoIS) as,
\begin{equation}
    \text{RoIS}(X, Y) = \frac{D(X, Y)}{d^*+\delta}
\end{equation}
where $\delta$ adjusts the smoothness of our metric. For our experiments, we use a normalized version of HSIC~\cite{hsic} in place of $D$, denoted by NHSIC.
\begin{equation}
    \text{NHSIC}(X, Y) = \frac{\text{HSIC}(X, Y)}{\sqrt{\text{HSIC}(X, X)\text{HSIC}(Y, Y)}}
\end{equation}

\section{Connecting spurious correlations through causal graphs\label{app:causal-spurious}}

Consider the causal graph provided in~\Cref{fig:causal-graph}. Due to the intervention on $B$, it becomes independent of its parent -- $A$. On the other hand, any change in $A$ must affect $B$. Therefore, a robust model cannot use features corresponding to $B$ to predict $A$. However, a bad model may be tempted to do so if, say, the features corresponding to $A$ are difficult to learn. Hence, a bad model is prone to show a drop in validation accuracy in predicting $A$ during interventions.

\section{Additional Results\label{app:additional-results}}

\subsection{Decision boundaries for easy-to-learn tasks\label{app:easy-circles-ratio-decision-boundaries}}

\begin{figure}[!ht]
    \centering
    \subcaptionbox{}{\includegraphics[width=0.31\textwidth]{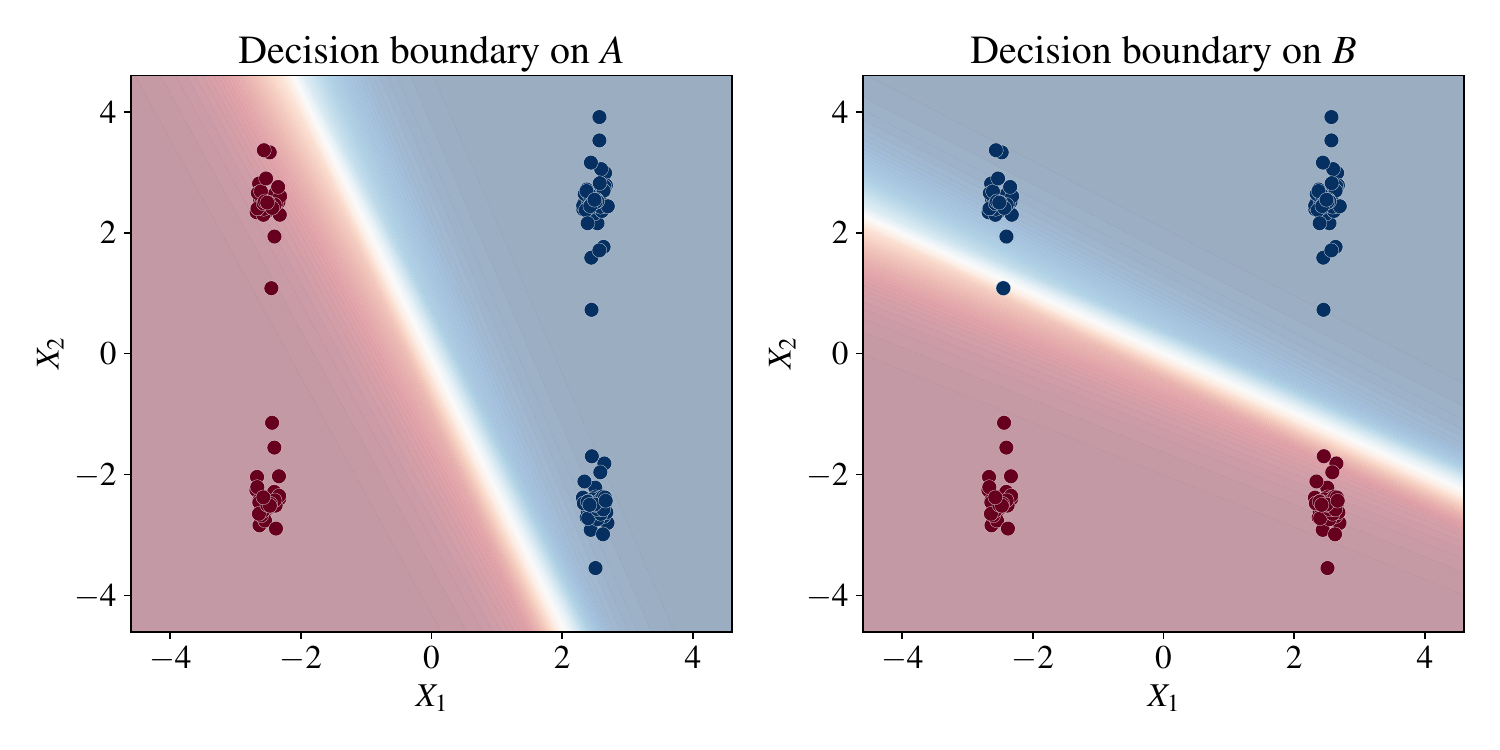}}
    \subcaptionbox{}{\includegraphics[width=0.31\textwidth]{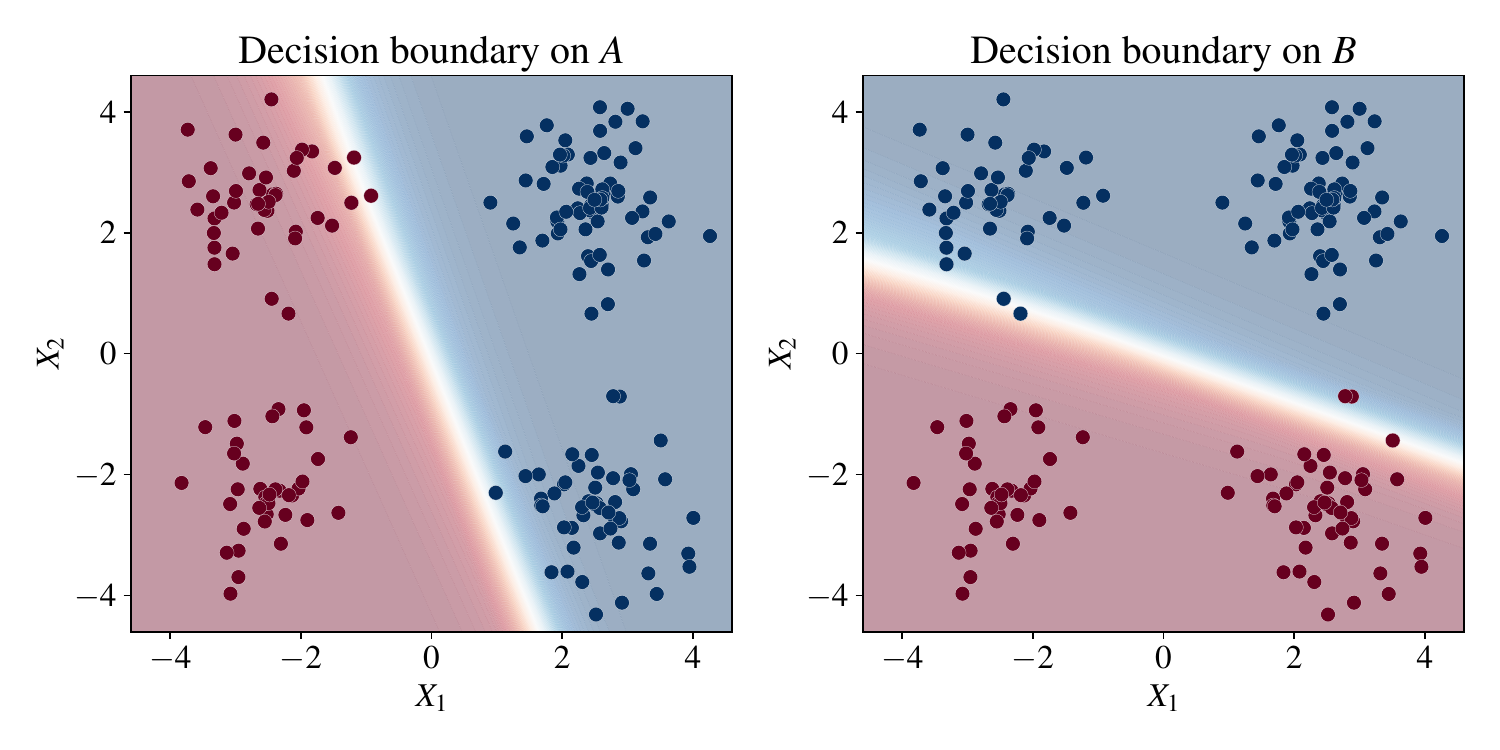}}
    \subcaptionbox{}{\includegraphics[width=0.31\textwidth]{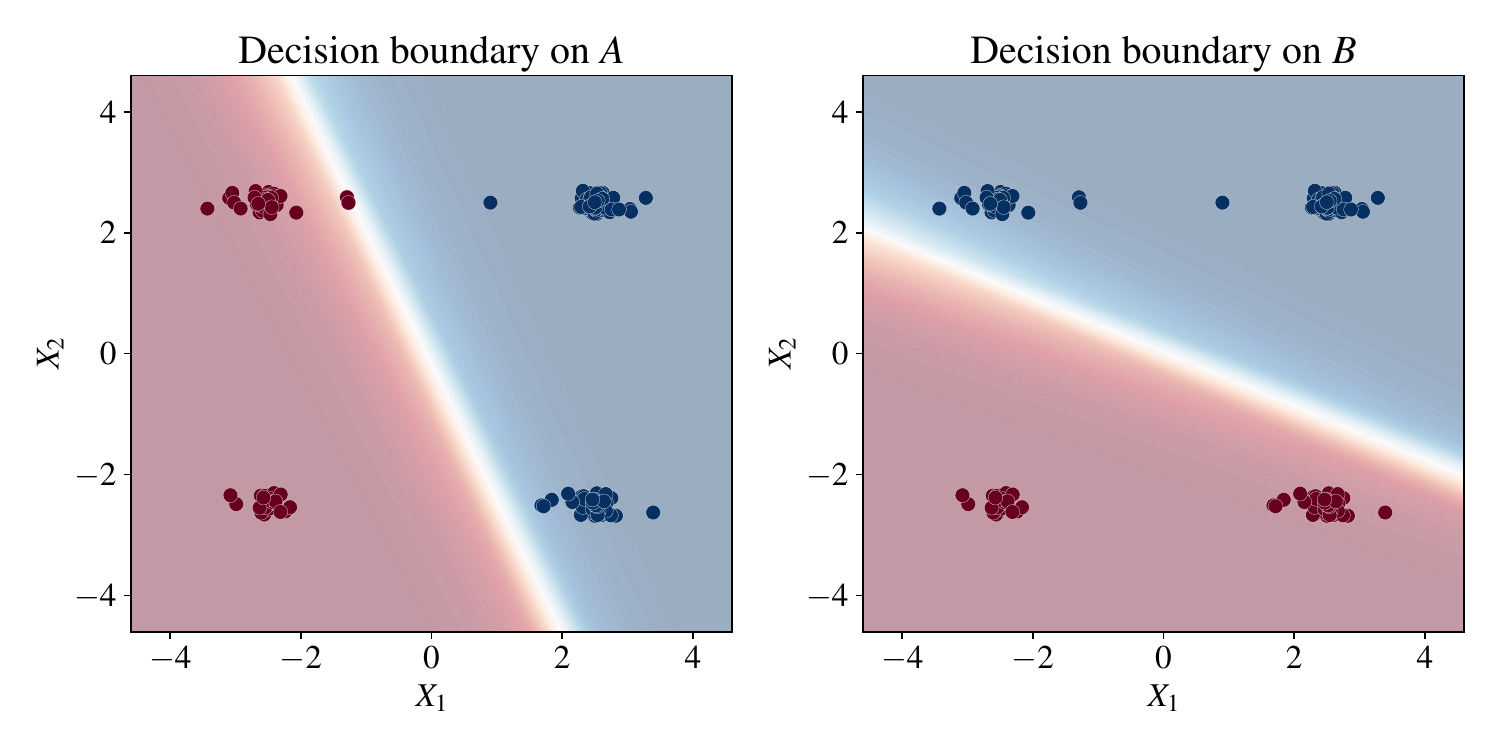}}
    \caption{Decision boundaries learned for different radius ratios}
    \label{fig:easy-circles-ratio-db}
\end{figure}

\end{document}